\documentclass[10pt,twocolumn,letterpaper]{article}

\usepackage[pagenumbers]{wacv} % To force page numbers, e.g. for an arXiv version

\definecolor{wacvblue}{rgb}{0.21,0.49,0.74}
\usepackage[pagebackref,breaklinks,colorlinks,allcolors=wacvblue]{hyperref}

\usepackage{graphicx}
\usepackage{booktabs}
\usepackage{multirow}

\usepackage{amsmath}
\usepackage{amssymb}
\usepackage{mathtools}
\usepackage{amsthm}

\title{ReBrain: Brain MRI Reconstruction from Sparse CT Slice via Retrieval-Augmented Diffusion}

\author{Junming Liu$^{1,2}$, Yifei Sun$^{2}$, Weihua Cheng$^{2}$, Yujin Kang$^{1}$, Yirong Chen$^{2}$, Ding Wang$^{2}$\thanks{Corresponding author.}, Guosun Zeng$^{1}$\footnotemark[1]\\
$^{1}$Tongji University\quad
$^{2}$Shanghai Artificial Intelligence Laboratory\\
{\tt\small liu\_junming6917@tongji.edu.cn\quad
wangding@pjlab.org.cn\quad
gszeng@tongji.edu.cn
}}

\begin{document}
\maketitle
\begin{abstract}
Magnetic Resonance Imaging (MRI) plays a crucial role in brain disease diagnosis, but it is not always feasible for certain patients due to physical or clinical constraints.
Recent studies attempt to synthesize MRI from Computed Tomography (CT) scans; however, low-dose protocols often result in highly sparse CT volumes with poor through-plane resolution, making accurate reconstruction of the full brain MRI volume particularly challenging.
To address this, we propose \textbf{ReBrain}, a retrieval-augmented diffusion framework for brain MRI reconstruction.
Given any 3D CT scan with limited slices, we first employ a Brownian Bridge Diffusion Model (BBDM) to synthesize MRI slices along the 2D dimension.
Simultaneously, we retrieve structurally and pathologically similar CT slices from a comprehensive prior database via a fine-tuned retrieval model.
These retrieved slices are used as references, incorporated through a ControlNet branch to guide the generation of intermediate MRI slices and ensure structural continuity.
We further account for rare retrieval failures when the database lacks suitable references and apply spherical linear interpolation to provide supplementary guidance.
Extensive experiments on SynthRAD2023 and BraTS demonstrate that ReBrain achieves state-of-the-art performance in cross-modal reconstruction under sparse conditions.
\end{abstract}    
\section{Introduction}
\label{sec:intro}

Brain imaging modalities such as Computed Tomography (CT) and Magnetic Resonance Imaging (MRI) are widely used in clinical diagnosis and neuroimaging analysis \cite{Hoffmann_2024_MRI, Ren_2024_MRI, Kim_2024_Controllable, Li_2024_Rethinking, Shamir_2025_Turorial}. Compared to CT, MRI typically incurs higher costs and longer acquisition times \cite{Ibrahim_2012_Cost, Mayerhoefer_2020_MRI, Reyes_2023_Cost}. Moreover, MRI is unsuitable for certain patient populations, such as those with implanted metallic devices \cite{Nazarian_2013_Circulation, Greene_2016_Considerations}. To address accessibility and availability issues, deep learning techniques have been proposed to reconstruct MRI images from CT scans \cite{Rahman_2023_Medical, Xing_2024_Cross, Bo_2025_FAMNet, Ha_2025_Multi}, enabling cross-modal synthesis.
However, clinical CT scans are often acquired with reduced sampling to minimize radiation exposure, resulting in thicker slices, lower resolution, and overall sparser volumes compared to conventional high-resolution MRI \cite{Sinsuat_2011_Influence, Guleng_2018_Radiation, Song_2024_I3Net}.
This sparsity poses a significant challenge for medical image tasks that involve reconstructing complete and high-fidelity MRI volumes from limited CT data.

\begin{figure}[t]
  \centering
  \fbox{\includegraphics[width=0.9\linewidth]{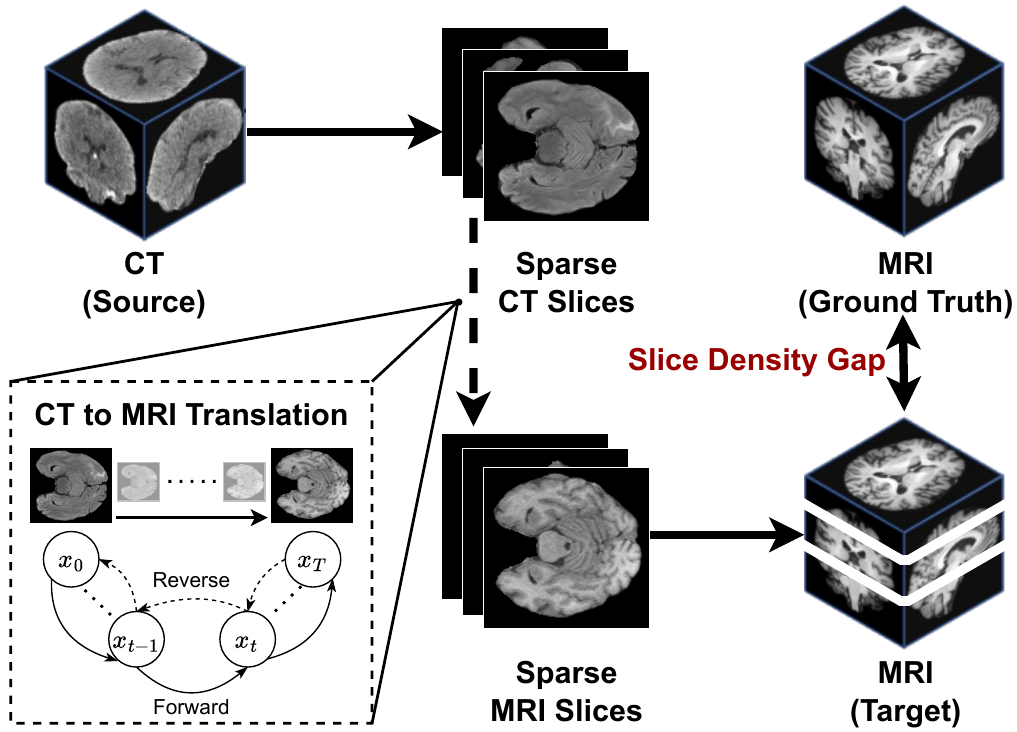}}
   \caption{Sparse CT slices (low axial resolution) lead to low-density MRI reconstructions with reduced inter-slice continuity, in contrast to the dense structure of standard MRI volumes.}
   \label{problem}
\end{figure}

Several approaches attempt to address the challenge of cross-modal reconstruction by synthesizing MRI directly from CT scans \cite{Ozbey_2023_Unsupervised, Choo_2024_CT2MRI, Kim_2024_ALDM}.
While these methods have shown promising results on datasets with densely sampled CT and MRI volumes, they often overlook the gap between ideal data conditions and real-world clinical acquisitions.
As a result, the reconstructed MRI volumes may lack sufficient volumetric continuity, thereby limiting their clinical utility when these methods are applied to sparse inputs.
To cope with such sparsity, other approaches employ generative models \cite{Goodfellow_2014_GAN, Ho_2020_Diffusion, xiao2025promptbasedadaptationlargescalevision} to interpolate within the source modality \cite{Zhu_2023_Make, Zhao_2024_DUE, Lee_2024_Reference}.
These methods have demonstrated strong generative capabilities, producing high-quality images with fine anatomical details that align well with clinical expectations \cite{Kazerouni_2023_Diffusion}.
However, they are restricted to within-modality synthesis, meaning that CT-based models must be trained and inferred on CT data to generate CT outputs, and the same applies to MRI. As such, they cannot be directly applied to cross-modal scenarios where only CT is available at inference time but the goal is to reconstruct MRI.

To address the limitations discussed above, we propose \textbf{ReBrain}, a 2D slice-based retrieval-augmented diffusion framework for cross-modal brain MRI reconstruction.
This is the first approach that jointly considers both CT sparsity and cross-modality generation, aiming to produce volumetrically continuous MRI volumes from sparse CT slices.

Inspired by prior work \cite{Choo_2024_CT2MRI, Lee_2025_EBDM}, we employ a Brownian Bridge Diffusion Model (BBDM) \cite{Li_2023_BBDM} to perform the CT-to-MRI generation.
Beyond outperforming deterministic solvers such as DDIM \cite{Song_2020_DDIM} and DPM-Solver \cite{Lu_2022_DPM-Solver} in preserving structural consistency, we introduce a directional noise formulation that explicitly guides generation from source to target. This leads to faster convergence and more coherent modality translation.

To address the limited continuity resulting from the sparse spatial distribution of CT slices, we introduce a retrieval-augmented strategy. Specifically, we first train a slice retrieval model capable of retrieving structurally similar brain slices based on the input CT slice.
We then construct a prior brain knowledge base, which stores a large collection of brain slices with diverse anatomical structures and spatial coverage.
For intermediate positions not represented in the input, structurally similar reference slices are retrieved from the knowledge base and provided as conditional inputs to a ControlNet \cite{Zhang_2023_Controlnet} module, which in turn guides the BBDM to synthesize context-aware MRI slices.
Compared to directly inserting retrieved slices as surrogates, our method yields substantially smoother and more continuous 3D reconstructions.
In scenarios where no sufficiently similar references can be retrieved, we employ an interpolation-based strategy using spherical linear interpolation (SLERP) \cite{Shoemake_1985_Slerp} between adjacent CT slices to control the MRI generation at the intermediate location, further improving volumetric consistency.

Our key contributions are as follows:
\begin{itemize}
    \item We present the first study that explicitly considers the low resolution commonly observed in clinical CT and reconstructs MRI from such realistically sparse inputs.
    \item We propose a novel retrieval-augmented diffusion approach using a BBDM to achieve structurally consistent cross-modal conversion.
    Rather than direct substitution, retrieved slices are employed as structural control signals to guide the generation process, producing smoother and more continuous volumetric reconstructions.
    \item We evaluate our method on the SynthRAD2023 \cite{Thummerer_2023_SynthRAD2023} and BraTS \cite{Menze_2015_BraTS} datasets, achieving state-of-the-art performance under sparse sampling conditions.
\end{itemize}
\section{Related Work}
\label{sec:r_w}

\subsection{Cross-Modal Medical Image Reconstruction}

Cross-modal medical image synthesis translates images between modalities to provide complementary diagnostic information and augment data \cite{Zhu_2017_CycleGAN, Zhou_2020_HiNet, Jaganathan_2023_Self}.
Medical imaging modalities can generally be categorized into X-ray, CT, and MRI, with acquisition cost and complexity increasing in that order \cite{Chen_2021_Anatomy, Hu_2025_CBCT}.
Early works focused on synthesizing higher-dimensional images from low-cost modalities.
For example, Liu \etal \cite{Liu_2024_DiffuX2CT} proposed DiffuX2CT, a conditional diffusion model that reconstructs 3D CT images from orthogonal biplanar X-rays, achieving structure-controllable and faithful CT synthesis. Similarly, Cai \etal \cite{Cai_2024_Structure} developed a structure-aware method for sparse-view X-ray 3D reconstruction, also targeting CT recovery from limited X-ray data.
Extending this line, recent studies have explored CT-to-MRI translation to leverage CT's accessibility while achieving MRI-like soft tissue contrast.
Choo \etal \cite{Choo_2024_CT2MRI} proposed a 2D BBDM with style key conditioning and inter-slice trajectory alignment to ensure slice-consistent 3D volumetric CT-to-MRI translation, addressing challenges of slice inconsistency in 2D diffusion models without the heavy cost of full 3D models.
Besides cross-modality translation, MRI modality conversion among contrasts is vital for comprehensive evaluation.
Kim and Park \cite{Kim_2024_ALDM} proposed an adaptive latent diffusion model with multiple switchable spatially adaptive normalization blocks for versatile 3D MRI image-to-image translation, supporting one-to-many modality synthesis with superior performance.
Most methods assume dense source modality data, but in practice, sources like CT may be sparsely sampled, yielding sparse target synthesis that may not meet clinical needs—especially for typically dense modalities like MRI. Bridging this gap remains an open challenge.

\subsection{Intra-Volume Slice Interpolation}

Sparse slice acquisition in CT and MRI often leads to discontinuous volumes, where large gaps between slices can result in missing anatomical information and reduced fidelity in downstream tasks. Intra-volume slice interpolation aims to densify such volumes by synthesizing intermediate slices, thereby improving resolution and spatial continuity while preserving clinically relevant structures.
SAINT \cite{Peng_2020_SAINT} introduces a spatially aware interpolation network that leverages voxel spacing to synthesize intermediate slices with adaptive resolution and low memory cost. Wu \etal \cite{Wu_2022_Slice} propose a multitask slice imputation method that enhances 3D isotropy by enforcing smoothness across all anatomical planes.
I3Net \cite{Song_2024_I3Net} exploits the high in-plane resolution of medical images through dual-branch interpolation and cross-view fusion, outperforming conventional super-resolution methods. DUE \cite{Zhao_2024_DUE} uses diffusion-based imputation to tackle sparsely annotated 3D data, demonstrating strong performance in volumetric reconstruction and uncertainty-aware explanation.
These methods effectively enhance volumetric continuity under sparse sampling conditions, though most are restricted to within-modality interpolation and are not designed for cross-modal reconstruction.
It is worth noting that slice sparsity differs from sparse-view acquisition \cite{Zang_2018_Super, Li_2023_Learning, Xie_2025_ACIND}. The former refers to large inter-slice spacing that leads to missing volumetric information, while the latter involves limited angular sampling during projection, resulting in degraded image quality. These are two distinct challenges.

\subsection{Diffusion Models in Medical Imaging}

Diffusion probabilistic models have emerged as a powerful alternative to GANs in medical image generation due to their superior ability to preserve fine anatomical details and avoid mode collapse \cite{Kazerouni_2023_Diffusion, liu2025fedrecon}. Their iterative, noise-aware generation process makes them particularly well-suited for applications requiring structural fidelity, such as reconstruction, synthesis, and segmentation \cite{Zhan_2024_MedM2G}.
MedSegDiff \cite{Wu_2024_MedSegDiff} is among the first to adapt diffusion models for general-purpose medical image segmentation, introducing dynamic conditional encoding and frequency-aware denoising to enhance accuracy across diverse modalities. Özbey \etal \cite{Ozbey_2023_Unsupervised} propose an adversarial diffusion framework for unsupervised cross-modal medical image translation, combining diffusion sampling with adversarial training to improve realism and alignment in unpaired settings.
Khader \etal \cite{Khader_2023_Denoising} demonstrate the effectiveness of denoising diffusion models for generating high-quality 3D CT and MRI volumes, with radiologist validation and downstream performance gains in segmentation tasks. Rahman \etal \cite{Rahman_2023_Ambiguous} leverage diffusion’s stochasticity to produce diverse segmentation outcomes, modeling inter-expert ambiguity in clinical scenarios more faithfully than deterministic networks.
These studies collectively highlight the versatility and effectiveness of diffusion models across a wide range of medical imaging tasks, setting a solid foundation for their application in more challenging settings such as cross-modal reconstruction.
\section{Methods}
\label{methods}

\begin{figure*}[t]
  \centering
  \includegraphics[width=0.95\linewidth]{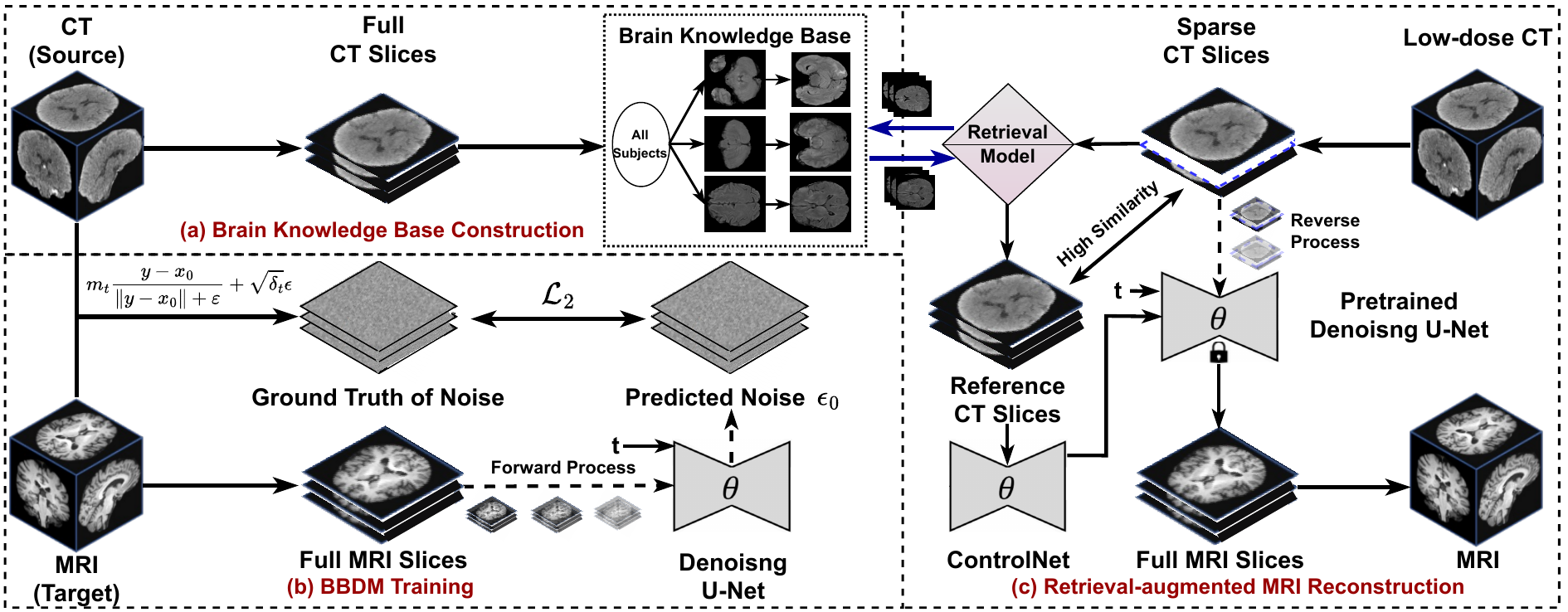}
   \caption{Overview of the \textbf{ReBrain} workflow.
   (a) We construct a brain knowledge base to enable effective retrieval. (b) We train a BBDM with optimized directional noise to better recover MRI from CT. (c) During inference, a fine-tuned retrieval model finds a similar reference slice, which is fed into the BBDM through a ControlNet branch to guide MRI reconstruction.}
   \label{workflow}
\end{figure*}

In this section, we introduce \textbf{ReBrain}, whose overall workflow is illustrated in Figure~\ref{workflow}. 
Designed in a 2D slice-based manner, our framework substantially lowers the computational cost compared to full 3D counterparts.

\subsection{Brownian Bridge Diffusion Model}
\label{bbdm}

We adopt the Brownian Bridge Diffusion Model (BBDM) \cite{Li_2023_BBDM, Choo_2024_CT2MRI} to model the transformation between source and target modalities. Instead of assuming a standard Gaussian prior as in traditional diffusion models, BBDM introduces a Brownian bridge formulation that interpolates between two arbitrary endpoints. Specifically, given a pair of images $(x, y)$, where $x$ is the target (e.g., MRI slice) and $y$ is the source (e.g., CT slice), the forward process is defined as:
\begin{equation}
q_{\text{BB}}(x_t | x_0, y) = \mathcal{N}(x_t; (1 - m_t)x_0 + m_t y,\ \delta_t I),
\end{equation}
where $x_0 = x$, $x_T = y$, $m_t = \frac{t}{T}$, and $\delta_t = 2s(m_t - m_t^2)$ controls the time-dependent variance, with $s$ being a scaling factor for sampling diversity.

The corresponding Markov transition kernel is given by:
\begin{equation}
q_{\text{BB}}(x_t | x_{t-1}, y) = \mathcal{N}(x_t; \mu_t,\ \delta_{t|t-1} I),
\end{equation}
where the variance update is $\delta_{t|t-1} = \delta_t - \delta_{t-1} \cdot \frac{(1 - m_t)^2}{(1 - m_{t-1})^2}$. The posterior distribution $q_{\text{BB}}(x_{t-1} | x_t, x_0, y)$ also has a closed-form Gaussian expression, with:
\begin{equation}
\tilde{\delta}_t = \frac{\delta_{t|t-1} \cdot \delta_{t-1}}{\delta_t}, \quad
\tilde{\mu}_t(x_t, x_0, y) = c_{xt} x_t + c_{yt} y + c_{\epsilon t} \cdot \tilde{\epsilon}_t,
\end{equation}
where $c_{xt}$, $c_{yt}$, and $c_{\epsilon t}$ are time-dependent coefficients.

The reverse process is parameterized as:
\begin{equation}
p_\theta(x_{t-1} | x_t, y) = \mathcal{N}(x_{t-1}; \mu_\theta(x_t, t),\ \tilde{\delta}_t I),
\end{equation}
where $\mu_\theta(x_t, t)$ is the predicted mean and shares the same variance $\tilde{\delta}_t$ as in the forward posterior.

To train the model, we minimize a simplified evidence lower bound (ELBO), where the objective is to match the posterior noise to the predicted noise:
\begin{equation}
\mathbb{E}_{x_0, t, y, \epsilon} \left[ \left\| \tilde{\epsilon}_t - \epsilon_\theta(x_t, t) \right\|_2^2 \right],
\end{equation}
and we propose a directional formulation for $\tilde{\epsilon}_t$, defined as:
\begin{equation}
\label{epsilon}
\tilde{\epsilon}_t = m_t \cdot \frac{y - x_0}{\|y - x_0\| + \varepsilon} + \sqrt{\delta_t} \cdot \epsilon,
\end{equation}
where $\epsilon \sim \mathcal{N}(0, I)$, $\varepsilon$ is a small constant to prevent division by zero and $\epsilon_\theta$ denotes the noise estimator parameterized by $\theta$.
This normalized direction term emphasizes structural guidance from source to target, leading to faster convergence and semantic consistency during training.

\subsection{Knowledge Retrieval}

Reconstructing a complete MRI volume from sparse CT slices can be approached in two ways.
The first is information closure, which infers MRI at unsampled positions from existing CT slices, typically via interpolation \cite{Baghaie_2014_Optimization, Peng_2020_SAINT,Wang_2023_Interpolating}.
The second is information supplementation, which leverages retrieval of additional relevant data to enrich and refine the reconstruction \cite{Pinapatruni_2022_Adversarial, Liu_2025_VaLiK}.
The former approach heavily depends on the quality and density of the available CT slices, often resulting in MRI reconstructions with unreliable textures. In contrast, the latter, under ideal conditions where the knowledge base contains sufficiently similar reference slices, provides more reliable guidance and produces reconstructions with richer and more realistic details, thus avoiding the artifacts typical of linear interpolation.

\begin{figure}[t]
  \centering
  \includegraphics[width=0.9\linewidth]{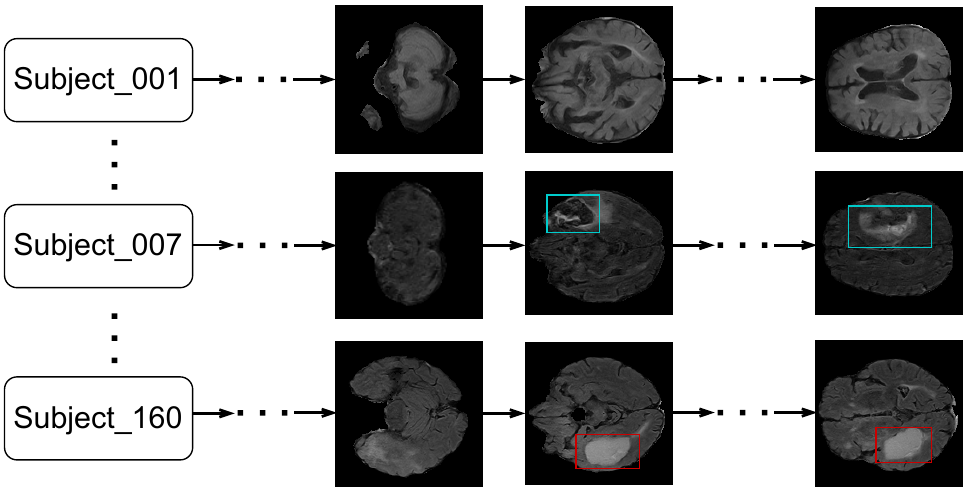}
   \caption{Illustration of the brain knowledge base structure.}
   \label{brain_kb}
\end{figure}

ReBrain adopts this retrieval-augmented strategy. Its performance upper bound in reconstructing complete MRI volumes is inherently limited by the prior knowledge stored in the database. Therefore, constructing a comprehensive and diverse knowledge base is crucial to increase the likelihood of successful retrieval, enabling structurally accurate control of the BBDM during MRI generation.
Given the structural and physiological similarities among human brains and the relatively small individual variations, it is generally feasible to build such a knowledge base using training data or additional homologous datasets. An illustration of the knowledge base structure is shown in Figure~\ref{brain_kb}.

Next, we describe how to train the retrieval model.
Since cross-subject similarity labels are not necessarily available, it is unclear which brain scans from different subjects are similar, so we adopt an intra-subject training strategy.
Specifically, through training on adjacent slices within each subject, the retrieval model \( \mathcal{R} \) learns to capture the shared anatomical structure of human brains, enabling it to retrieve similar slices across different subjects.
To implement this, we use a pretrained Vision Transformer (ViT) \cite{Dosovitskiy_2020_ViT} and fine-tune it using a contrastive learning loss.
We consider each brain volume as multiple slices with defined spatial positions.
For any slice at position $p$, we compute the similarity between its embedding and those of slices within a local neighborhood, specifically at positions $p \pm k$. \( \mathcal{R} \) is trained to assign higher similarity scores to slices close in position, reflecting anatomical continuity.
Initially, the similarity computation was performed bidirectionally, considering both $p - k$ and $p + k$ relative to $p$. However, we observed redundancy: the similarity between $p$ and $p - k$ is effectively optimized when processing position $p - k$ against $p$. Therefore, a unidirectional contrastive loss suffices, reducing complexity while maintaining performance.

Let \(\phi(\cdot)\) denote the encoder within \(\mathcal{R}\) that maps each brain slice to an embedding used for contrastive learning.
For a given slice $x_p$ at position $p$, we define its positive set $\mathcal{P}_p = \{x_{p+k} \mid k \in \mathcal{K} \}$, where $\mathcal{K}$ is a set of predefined positional offsets (e.g., $\mathcal{K} = \{1,2,\dots,K\}$). The contrastive loss encourages the embeddings of $x_p$ and its neighboring slices to be close, while pushing apart embeddings from slices sampled across the dataset as negatives.
The similarity between two slices is defined as:
\begin{equation}
\operatorname{sim}(x_i, x_j) = \exp\left( \frac{\phi(x_i)^\top \phi(x_j)}{\alpha} \right),
\end{equation}
where $\alpha$ is a temperature hyperparameter.
The contrastive loss is given by:
\begin{equation}
\mathcal{L}_\text{contrast} = - \log \frac{
\sum\limits_{x^+ \in \mathcal{P}_p} \operatorname{sim}(x_p, x^+)
}{
\sum\limits_{x^- \in \mathcal{N}_p} \operatorname{sim}(x_p, x^-) +
\sum\limits_{x^+ \in \mathcal{P}_p} \operatorname{sim}(x_p, x^+)
},
\end{equation}
where $\mathcal{N}_p$ denotes the negative set, typically composed of randomly sampled slices from other positions or subjects within the batch.
This loss encourages \(\phi(\cdot)\) to map anatomically adjacent slices to nearby points in the embedding space, enabling structure-aware retrieval during inference.

To further enforce structural similarity, we incorporate a perceptual contrastive loss based on intermediate features extracted from $\phi(\cdot)$.
This encourages not only the final embeddings but also the structural features within the encoder to be aligned.
We define the distance function:
\begin{equation}
d(x_i, x_j) = \exp\left( - \frac{\|\phi(x_i) - \phi(x_j)\|_2^2}{\beta} \right),
\end{equation}
where $\beta$ controls sensitivity in the perceptual feature space.
The perceptual contrastive loss is defined as:
\begin{equation}
\mathcal{L}_\text{percept} = - \log \frac{
\sum\limits_{x^+ \in \mathcal{P}_p} d(x_p, x^+)
}{
\sum\limits_{x^- \in \mathcal{N}_p} d(x_p, x^-)
+ \sum\limits_{x^+ \in \mathcal{P}_p} d(x_p, x^+)
}.
\end{equation}

The total loss combines both terms with a balancing hyperparameter $\lambda$:
\begin{equation}
\mathcal{L} = \mathcal{L}_\text{contrast} + \lambda \mathcal{L}_\text{percept}.
\end{equation}
This joint optimization aligns both semantic embeddings and perceptual structures, improving retrieval for structure-aware guidance.

\subsection{Conditional Guidance in BBDM via ControlNet}

In the following, we describe how the retrieved slice is incorporated into ControlNet to guide BBDM generation.
During training, the original BBDM backbone parameters $\theta_{\text{BB}}$ are frozen, while ControlNet parameters $\theta_{\text{CN}}$ are optimized to leverage the reference $r$ and the source CT slice \( y \).
We do not require additional structure extraction from $r$ and instead directly use it as the condition\footnote{This step is inspired by the implementation in \href{https://github.com/Warvito/generative_brain_controlnet}{Gtihub}.}.
Note that in this setting, the reconstruction target $x_0$ is the MRI paired with $r$, rather than the MRI originally paired with $y$.
For each ControlNet block, the output is computed as
\begin{equation}
h = F(x; \Theta) + Z\big( F\big( x + Z(r; \Theta_{z1}); \Theta_c \big); \Theta_{z2} \big),
\label{eq:controlnet_block}
\end{equation}
where \( x \) is the input to the block, \( r \) is the conditional input, \( F(\cdot; \cdot) \) denotes a neural network block parameterized by \( \Theta \), and \( Z(\cdot; \cdot) \) is a zero-initialized 1×1 convolution layer with trainable parameters \( \Theta_{z1} \), \( \Theta_c \), and \( \Theta_{z2} \). This residual structure enables ControlNet to inject conditional features progressively while preserving the pretrained backbone.

The training objective minimizes the simplified ELBO loss:
\begin{equation}
\mathcal{L}_{\text{CN}} = \mathbb{E}_{x_0, t, y, \epsilon, r} \left[ \left\| \tilde{\epsilon}_t - \epsilon_\theta(x_t, t, r) \right\|_2^2 \right],
\end{equation}
where \( x_t \) is the noisy latent at time \( t \) generated via the forward process of BBDM with the terminal condition \( x_T = y \), and \( \tilde{\epsilon}_t \) is the directional noise defined in Equation~\eqref{epsilon}.

At inference time, given an input source $y$ and retrieved reference $r$, the reverse diffusion sampling is performed iteratively:
\begin{equation}
x_{t-1} \sim p_\theta(x_{t-1} | x_t, y, r) = \mathcal{N}(x_{t-1}; \mu_\theta(x_t, t, r), \tilde{\delta}_t I),
\end{equation}
where the mean \( \mu_\theta \) is computed using the augmented noise predictor \( \epsilon_\theta(x_t, t, r) \) as above.
Using \( y \) as the primary input while employing the retrieved neighbor \( r \) for structural control leads to smoother and more coherent generation than directly using \( y = r \). This is because replacing \( y \) with \( r \) discards the semantic continuity provided by the original source, forcing the model to rely solely on the potentially noisy or mismatched reference. In contrast, preserving \( y \) allows the model to maintain contextual consistency while softly adapting structural cues from \( r \), resulting in more stable and anatomically plausible outputs.

\subsection{Latent Interpolation as a Fallback for Retrieval}

We consider the retrieval failed if the highest cosine similarity falls below a threshold \(\tau\), which is set as the average of the top 5\% cosine similarities across all queries in the training set.
In such cases, we adopt an interpolation strategy to generate control signals when no suitable reference slice is found.
This fallback guarantees stable outputs even in rare cases where the database lacks relevant references.
Specifically, given two adjacent slices \( x_i \) and \( x_j \) surrounding the intermediate slice, we perform spherical linear interpolation (slerp) \cite{Shoemake_1985_Slerp, Wang_2023_Interpolating} in the latent or feature space to synthesize an intermediate control input.

Formally, the interpolated slice \( x_\alpha \) is computed as:
\begin{equation}
x_\alpha = \mathrm{slerp}(x_i, x_j; \alpha) = \frac{\sin((1-\alpha)\theta)}{\sin \theta} x_i + \frac{\sin(\alpha \theta)}{\sin \theta} x_j,
\end{equation}
where \( \alpha \in (0,1) \) determines the interpolation ratio, and \( \theta = \arccos\left( \frac{x_i^\top x_j}{\|x_i\| \|x_j\|} \right) \) is the angle between the two vectors.

This interpolation scheme provides geometrically meaningful transitions between neighboring slices, offering a plausible structural prior that acts as a fallback to compensate retrieval failures and preserve structural continuity.
\section{Experiments}

\begin{table*}[ht]
\centering
\begin{tabular}{lcccccccc}
\toprule
\multirow{2}{*}{Method} & \multicolumn{4}{c}{SynthRAD2023} & \multicolumn{4}{c}{BraTS} \\
\cmidrule(lr){2-5} \cmidrule(lr){6-9}
& NRMSE$\downarrow$ & PSNR$\uparrow$ & SSIM$\uparrow$ & I-SSIM$\uparrow$
& NRMSE$\downarrow$ & PSNR$\uparrow$ & SSIM$\uparrow$ & I-SSIM$\uparrow$ \\
\midrule
MaskGAN      & 0.1526 & 17.685 & 0.5883 & 0.9093 & 0.1035 & 20.171 & 0.7028 & 0.9215 \\
CT2MR        & 0.1138 & 19.293 & 0.6015 & 0.9484 & 0.0643 & 23.844 & 0.8535 & 0.9489 \\
Dual-approx Bridge & 0.1093 & 18.178 & 0.6162 & 0.9382 & 0.0706 & 24.510 & 0.8419 & 0.9404 \\
I3Net        & 0.1653 & 11.775 & 0.5581 & 0.9407 & 0.1502 & 18.914 & 0.8074 & 0.9530 \\
ALDM       & 0.1336 & 17.561 & 0.6266 & 0.9368 & 0.0808 & 21.921 & 0.7930 & 0.9544 \\
3D-mDAUNet        & 0.1069 & 18.662 & 0.6331 & 0.9466 & 0.0602 & 24.646 & 0.8378 & \textbf{0.9597} \\
\textbf{ReBrain (Ours)} & \textbf{0.1054} & \textbf{20.759} & \textbf{0.6682} & \textbf{0.9662} & \textbf{0.0551} & \textbf{25.607} & \textbf{0.8887} & 0.9575 \\
\bottomrule
\end{tabular}
\caption{Quantitative comparison under half-resolution input on SynthRAD2023 and BraTS datasets. Results are averaged over three seeds, with observed variations being minimal; more comprehensive results including standard deviations are provided in Appendix~D.}
\label{tab:quantitative_comparison}
\end{table*}

\begin{figure*}[t]
  \centering
  \includegraphics[width=1.0\linewidth]{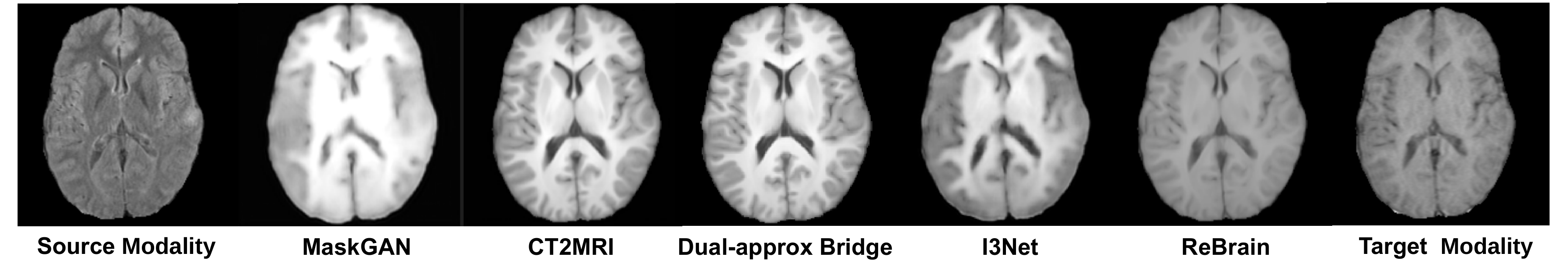}
    \caption{Representative cross-modal reconstruction results on the BraTS dataset. More results are provided in Appendix~D.}
   \label{Cross_BraTS}
\end{figure*}

\subsection{Experimental Setups}

\textbf{Datasets.} We evaluate our method on two publicly available datasets: SynthRAD2023 \cite{Thummerer_2023_SynthRAD2023} and BraTS \cite{Menze_2015_BraTS}.
\textbf{SynthRAD2023} is a multi-institutional benchmark dataset for synthetic CT generation in radiotherapy. It includes paired CT, CBCT, and MRI scans from 540 patients across different anatomical regions. We use only the brain subset, which provides rigidly registered T1-weighted MRI and CT images, enabling accurate evaluation of cross-modality generation models under controlled conditions.
\textbf{BraTS} provides multimodal MRI scans (FLAIR, T1, T1ce, T2) with pixel-level tumor annotations. Although it does not include CT data, it offers a realistic benchmark for testing model performance on incomplete modalities in clinical brain tumor scenarios.
In our experiments, we use FLAIR MRIs as a substitute for CT to reconstruct T1-weighted MRIs.

\textbf{Baselines.}
For comparison, we evaluated several 2D and 3D models for image-to-image translation (I2I), generally using their default settings unless otherwise noted. These include MaskGAN \cite{Phan_2023_MaskGAN}, a cyclic GAN originally designed for unpaired 2D medical I2I and adapted here for paired training;
ALDM \cite{Kim_2024_ALDM}, which implements a 3D latent diffusion framework;
I3Net \cite{Song_2024_I3Net}, an inter-slice interpolation network for 3D medical image super-resolution.
CT2MRI \cite{Choo_2024_CT2MRI}, a 2D BBDM for 3D CT-to-MRI translation with style-consistent and slice-aligned generation; and Dual-approx Bridge \cite{Xiao_2025_Dual-approx_Bridge}, a BBDM with dual networks that achieves faithful translations; 
and 3D-mDAUNet \cite{Ha_2025_Multi}, a multi-resolution guided 3D GAN for image translation.

\textbf{Implementation.}
For each dataset, we constructed the retrieval database using its full training set, and trained our method as well as all baselines on the complete training data.
The hyperparameters \(\alpha\) and \(\beta\) were set to 1, while \(\lambda\) was set to 0.5.
For the validation or test sets, we simulated slice sparsity by uniformly reducing the number of slices along the axial axis. For example, reducing 200 slices to 100 corresponds to $\times$2 lower axial resolution, effectively mimicking the outcome of thicker-slice acquisition across the same volume extent.
Following CT2MRI \cite{Choo_2024_CT2MRI}, all 2D models operated on axial slices, and all diffusion models were trained with 1000 diffusion steps and sampled using 100 DDIM steps. All models were trained for 100{,}000 iterations on a single NVIDIA A100 GPU with 80GB memory.

\textbf{Evaluation Metrics} 
We evaluate the reconstructed MRI volumes using three standard image quality metrics: normalized root mean square error (NRMSE), peak signal-to-noise ratio (PSNR), and structural similarity index measure (SSIM).
To assess reconstruction continuity along the axial direction, we report Intra-Slice SSIM (I-SSIM), the average SSIM between adjacent reconstructed slices.
For methods that disregard slice sparsity and do not use interpolation, we approximate intermediate slices by summing adjacent reconstructed slices for comparison with dense ground-truth. Methods handling sparsity typically use slice interpolation, trained on dense MRI and reconstructing MRI from sparse CT during inference.
We conducted experiments over three random seeds and report the mean values.

\subsection{Results}

We evaluated our method on the SynthRAD2023 and BraTS datasets. Quantitative comparisons under half-resolution input are summarized in Table~\ref{tab:quantitative_comparison}. 
Under the low-resolution setting, all methods that do not explicitly address slice sparsity perform poorly in reconstruction, as they fail to infer intermediate MRI slices from sparse CT inputs.
Although I3Net performs intermediate slice reconstruction, its reliance on CT-based inference without access to true MRI information leads to severely degraded and often catastrophic results.
In contrast, ReBrain produces high-quality reconstructed slices and achieves state-of-the-art performance across all four evaluation metrics on both datasets.

We selected a representative example of cross-modal reconstruction on BraTS for each method, shown in Figure~\ref{Cross_BraTS}. ReBrain, leveraging optimized noise directions, converges faster than other BBDM-based approaches such as CT2MRI and Dual-approx Bridge, while delivering highly competitive image reconstruction quality.

\begin{figure}[t]
  \centering
  \includegraphics[width=1.0\linewidth]{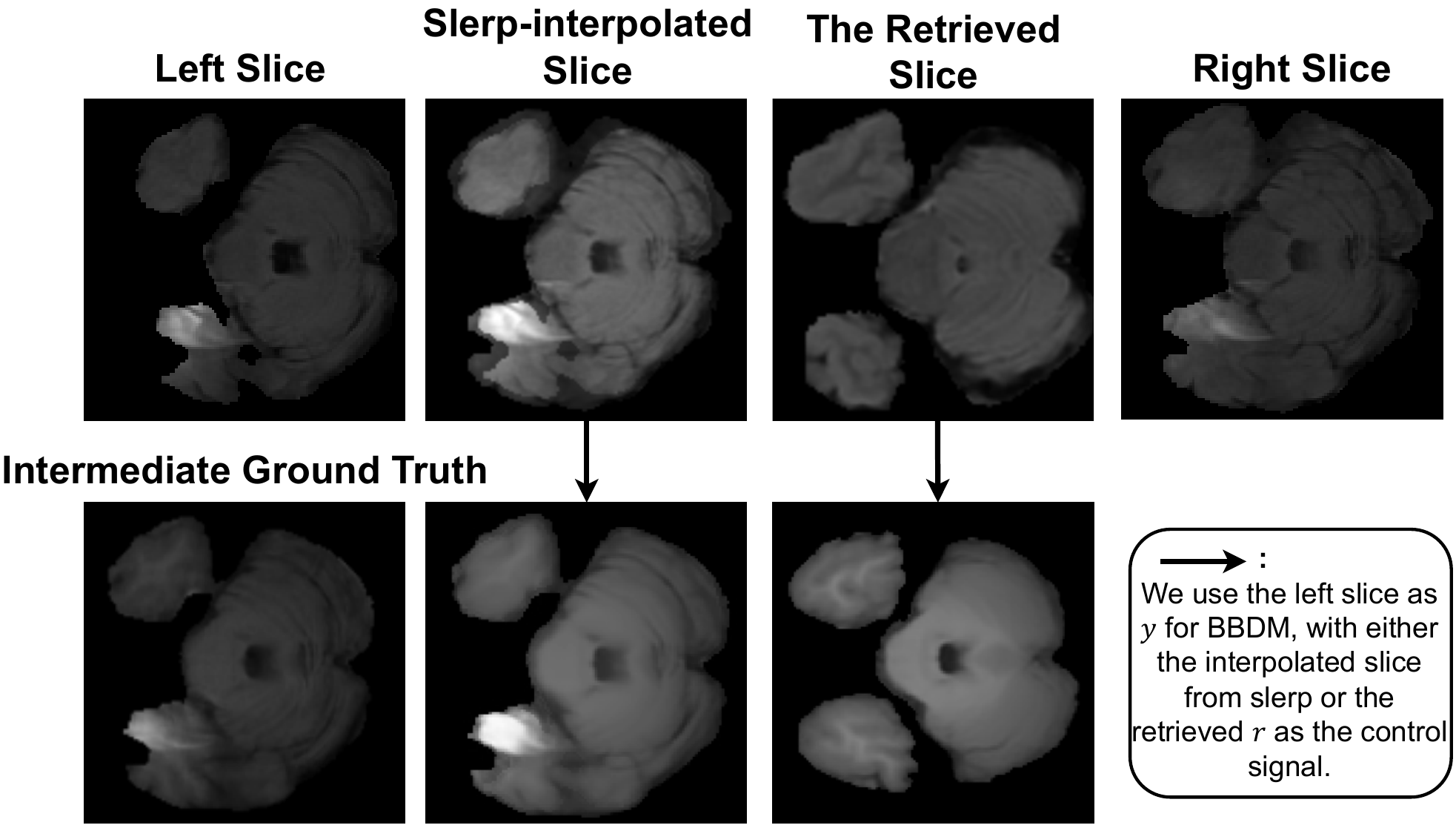}
  \caption{
  Comparison of generation results using interpolation and retrieval under varying similarity conditions on BraTS.
  }
  \label{fig:slerp_comparison}
\end{figure}

We further present the generation results of \textit{slerp} in Figure~\ref{fig:slerp_comparison}. As shown, when the similarity between the target slice \( y \) and the retrieved slice \( r \) is low (\( \text{sim}(y, r) < \tau \)), directly using \( r \) for reconstruction leads to inferior results. In contrast, the interpolated slice from \textit{slerp} yields more faithful outputs under this condition.

\subsection{Ablation Study}

In this subsection, we present a comprehensive ablation study to demonstrate the effectiveness of each component in ReBrain, with a particular focus on validating the reliability of our retrieval-augmented strategy.

\begin{table}[t]
\centering
\setlength{\tabcolsep}{2pt}
\scalebox{0.95}{
\begin{tabular}{lccccc}
\toprule
Method & NRMSE$\downarrow$ & PSNR$\uparrow$ & SSIM$\uparrow$ & I-SSIM$\uparrow$ & Time$\downarrow$ \\
\midrule
Baseline BBDM & 0.1239 & 19.103 & 0.6427 & 0.9558 & 36.61h \\
\textbf{+ Noise Opt.} & \textbf{0.1054} & \textbf{20.759} & \textbf{0.6682} & \textbf{0.9662} & \textbf{14.25h} \\
\bottomrule
\end{tabular}
}
\caption{Effect of noise optimization in BBDM on the SynthRAD2023 dataset.
}
\label{tab:noise_optimization}
\end{table}

\textbf{Effect of Noise Optimization in BBDM.}  
As shown in Table~\ref{tab:noise_optimization}, by optimizing the noise direction in BBDM, we achieve notable improvements over the baseline, with a PSNR increase of 1.656\% and an SSIM gain of 0.0255\%.
Additionally, the training time is reduced by approximately 60\%, with the model converging after about 40\% of the original number of iterations.

\begin{table}[t]
\centering
\setlength{\tabcolsep}{2.4pt}
\scalebox{0.95}{
\begin{tabular}{lccccc}
\toprule
Model & Acc.$\uparrow$ & NRMSE$\downarrow$ & PSNR$\uparrow$ & SSIM$\uparrow$ & I-SSIM$\uparrow$ \\
\midrule
Pretrained ViT     & 80.46\% & 0.0784 & 22.617 & 0.8127 & 0.9410 \\
+\(\mathcal{L}_\text{contrast}\)        & 94.54\% & 0.0601 & 24.667 & 0.8699 & 0.9462 \\
\textbf{+\(\mathcal{L}_\text{percept}\)} & \textbf{95.96\%} & \textbf{0.0551} & \textbf{25.607} & \textbf{0.8887} & \textbf{0.9575} \\
\bottomrule
\end{tabular}
}
\caption{Retrieval accuracy and corresponding reconstruction performance using ViT-based retrieval models with ablated loss functions on the BraTS dataset.}
\label{tab:retrieval_model}
\end{table}

\textbf{Effect of Retrieval Model.}
To better evaluate the effectiveness of the retrieval module, we introduce a retrieval accuracy metric on the BraTS dataset. Specifically, we fine-tune a pretrained ViT on the validation set and partition it into two equal parts: one as queries (images used to search the database) and the other as the retrieval database (images used to find matches for the queries). Retrieval accuracy is measured by the model’s ability to correctly match the subject ID of each query slice to the subject ID of the top-1 retrieved slice.
As shown in Table~\ref{tab:retrieval_model}, we compare three variants of the retrieval model, showing their retrieval accuracy and corresponding volume reconstruction performance.

\begin{table}[t]
\centering
\setlength{\tabcolsep}{2.8pt}
\scalebox{0.95}{
\begin{tabular}{lcccc}
\toprule
Setting & NRMSE$\downarrow$ & PSNR$\uparrow$ & SSIM$\uparrow$ & I-SSIM$\uparrow$ \\
\midrule
w/o ControlNet & 0.5939 & 24.571 & 0.8460 & 0.9237 \\
\textbf{w/ ControlNet (Ours)} & \textbf{0.0551} & \textbf{25.607} & \textbf{0.8887} & \textbf{0.9575} \\
\bottomrule
\end{tabular}
}
\caption{Effect of ControlNet on BraTS.
}
\label{tab:controlnet_effect}
\end{table}

\textbf{Effect of ControlNet.}  
As shown in Table~\ref{tab:controlnet_effect}, omitting the ControlNet and directly feeding the retrieved slices as input leads to a drop in performance. This is because the retrieved slices, while similar, do not originate from the same patient and thus can only serve as a rough reference rather than authentic data for the target case. By using the sparse CT slice \( y \) as the primary input to BBDM and the retrieved slice \( r \) as a control signal, ControlNet effectively integrates both sources of information. In essence, \( r \) guides \( y \) in generating a more accurate MRI, which is also more consistent with ethical considerations in medical image synthesis.

\subsection{Further Analysis}

\begin{figure}[t]
    \centering
    \includegraphics[width=0.9\linewidth]{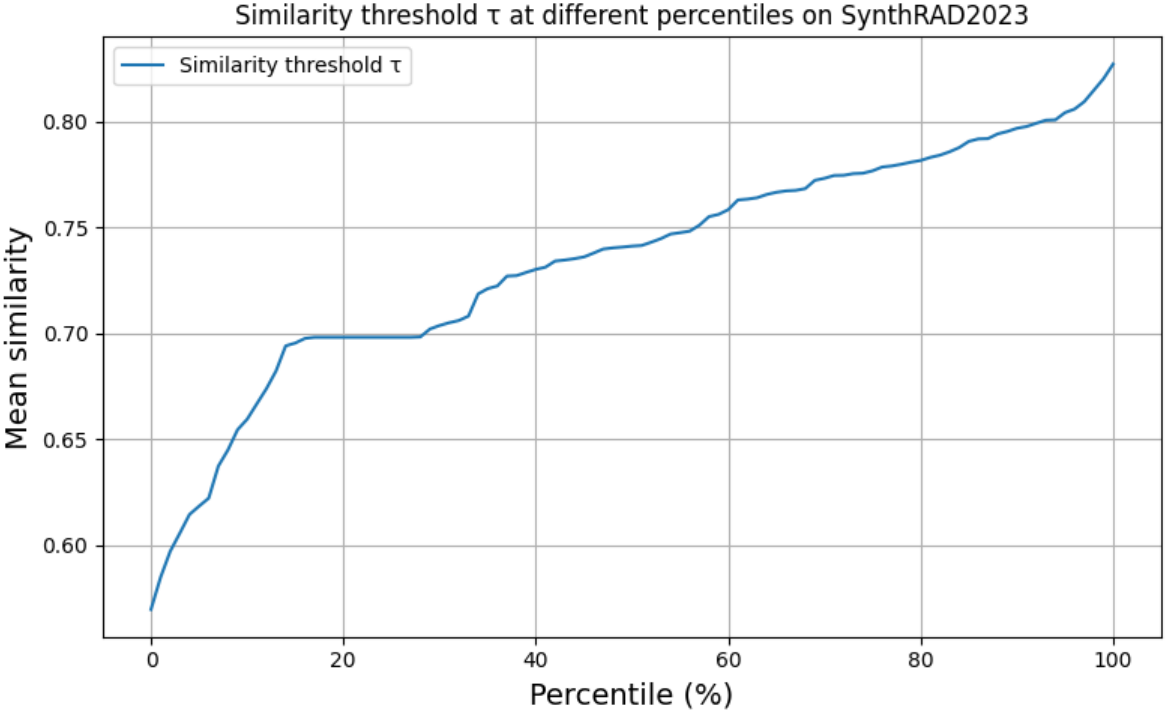}
    \caption{Threshold \(\tau\) values at different percentiles of similarity scores in the SynthRAD2023 training set.}
    \label{fig:tau_distribution}
\end{figure}

\textbf{Selection of the Similarity Threshold \(\tau\).}
We investigate the selection of threshold \(\tau\) used to determine retrieval success. \(\tau\) is empirically chosen from the training set by computing pairwise similarities of random CT slices. The threshold corresponds to the top-\(p\%\) percentile of similarity scores, with \(p=5\) by default to ensure informative references.
Figure~\ref{fig:tau_distribution} shows \(\tau\) values at different percentiles on SynthRAD2023. Higher percentiles impose stricter similarity, resulting in larger \(\tau\), guiding the trade-off between retrieval coverage and quality.

\textbf{Selection of Retrieval Database Size.}  
To evaluate the sensitivity to the retrieval database, we conduct experiments using subsets of the training set as the retrieval database. To control for variables, the retrieval model itself is always trained on the full training set. Results on BraTS are reported in Table~\ref{database}. As the database size decreases, retrieval success rate drops accordingly; however, our SLERP-based interpolation effectively compensates for poorly matched slices, preventing hallucinations in the reconstructed MRI. Even with only 30\% of the training set, retrieval success remains above 40\%, indicating robustness. This is reasonable since central brain slices are structurally similar, providing a reliable knowledge reference, and the generation is guided both by retrieved slices \( r \) and the original input \( y \).

\begin{table}[t]
    \centering
    \renewcommand{\arraystretch}{0.9}
    \setlength{\tabcolsep}{3pt}
    \begin{tabular}{lccccc}
        \toprule
        Size & Rate (\%) & NRMSE↓ & PSNR↑ & SSIM↑ & I-SSIM↑ \\
        \midrule
        100\% & 95.96 & 0.0551 & 25.607 & 0.8887 & 0.9575 \\
        70\%  & 84.49  & 0.0605 & 25.223 & 0.8721 & 0.9546 \\
        50\%  & 63.62  & 0.0672 & 23.892 & 0.8520 & 0.9416 \\
        30\%  & 42.50  & 0.0755 & 23.103 & 0.8605 & 0.9384 \\
        \bottomrule
    \end{tabular}
    \caption{Effect of database size on BraTS.}
    \label{database}
\end{table}

\begin{figure}[t]
    \centering
    \includegraphics[width=0.47\textwidth]{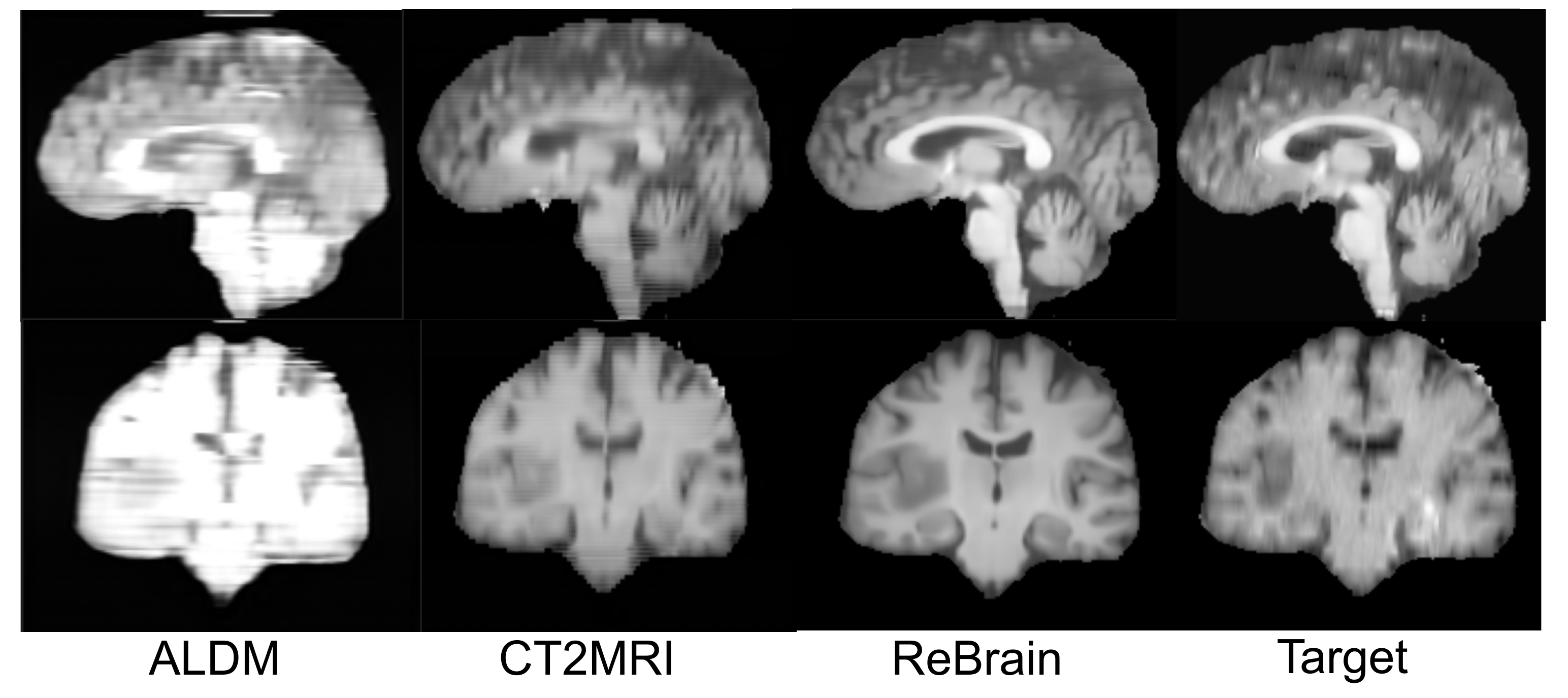}
    \caption{Coronal and sagittal views across different methods.}
    \label{fig:3DI}
\end{figure}

\textbf{3D Volume Generation and Multi-View Evaluation.}  
To assess volumetric consistency, we evaluate coronal and sagittal views across methods. Figure~\ref{fig:3DI} shows representative reconstructions, highlighting that ReBrain achieves the most coherent cross-slice generation. Additional analyses and examples are provided in Appendix~D.

\textbf{Impact of Retrieval Similarity on Generation.}  
We provide a detailed analysis in Appendix~D, showing how the similarity between input slice \(y\) and retrieved slice \(r\) affects generation, with ReBrain demonstrating superior robustness and stability compared to baselines.

\section{Limitations}

\textbf{Ethical and practical considerations.}  
Our framework uses previously observed, structurally similar slices only as a conditional reference to guide MRI reconstruction, rather than directly copying prior patient data. This ensures that retrieved slices act as a "soft hint," not a precise replication.
While maintaining reasonable 3D structural continuity is one of the aims, it is achieved as part of balancing multiple reconstruction objectives rather than being the sole focus.
SLERP interpolation is applied when retrieval fails or yields low-similarity slices, serving as a safe fallback that minimizes hallucinated structures.

Importantly, in clinical scenarios where the original CT lacks complete pathological information, no reconstruction method can invent missing lesions. If a retrieved reference contains relevant pathology, it can assist in highlighting critical structures, potentially benefiting the patient. Conversely, if the patient is healthy or the reference has no pathology, any introduced errors are minor and limited to a careful review. Overall, this strategy substantially reduces hallucinations while keeping clinical risk very low, with potential benefits outweighing the unlikely costs. Detailed discussion is provided in Appendix~E.
\section{Conclusion}
In this work, we present ReBrain, a retrieval-augmented diffusion framework for reconstructing brain MRI from sparse CT inputs. Unlike prior methods that assume dense sampling or perform within-modality synthesis, ReBrain addresses both the modality gap and data sparsity. By combining BBDM with a retrieval-based ControlNet, it leverages local CT context and non-local anatomical priors to generate structurally coherent MRI volumes.
Experiments show that ReBrain produces high-fidelity reconstructions under sparse conditions, offering a practical and accurate approach for clinically constrained settings.

\section{Acknowledgments}

The research was supported by Shanghai Artificial Intelligence Laboratory, the National Key R\&D Program of China (Grant No. 2022ZD0160201) and the Science and Technology Commission of Shanghai Municipality (Grant No. 22DZ1100102).

{
    \small
    \bibliographystyle{ieeenat_fullname}
    \bibliography{main}
}

\clearpage

\appendix
\section{Datasets}

\begin{figure}[t]
  \centering
  \includegraphics[width=\linewidth]{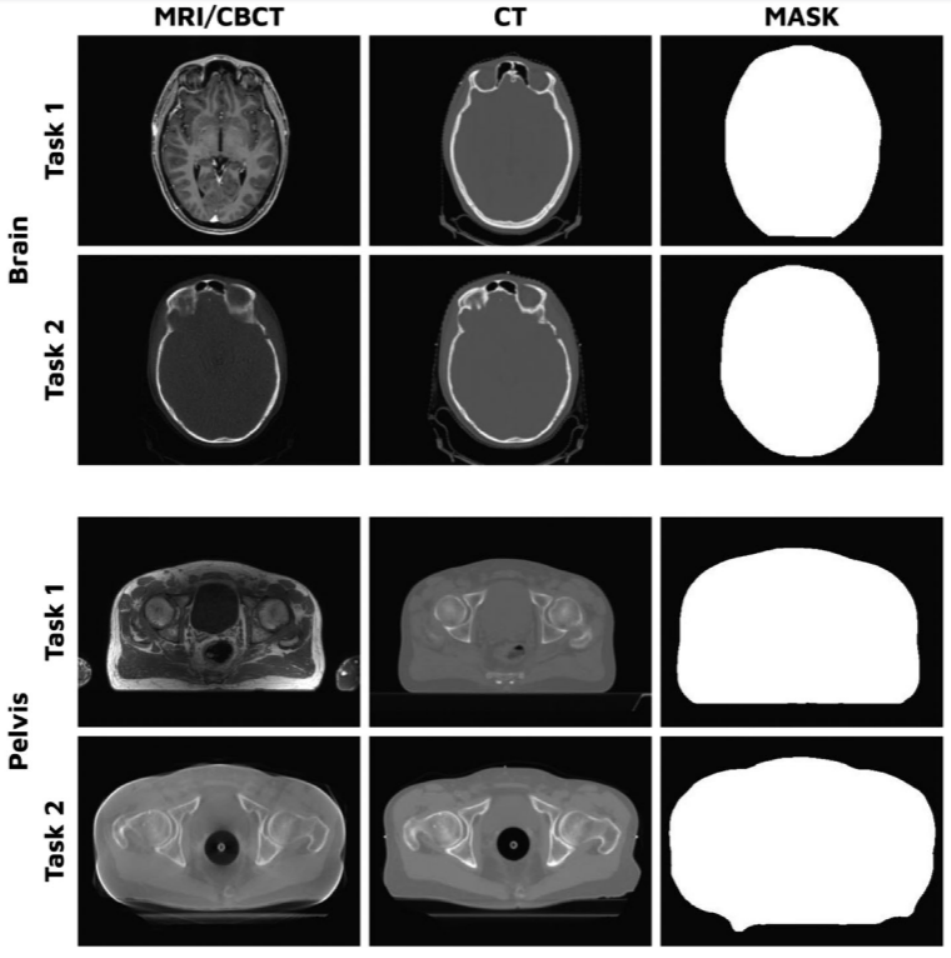}
  \caption{Example images for all tasks and anatomies in the synthRAD2023 dataset. The top row shows images for task 1 brain, the middle-top for task 1 pelvis, the middle-bottom for task 2 brain, and the bottom for task 2 pelvis. The first column shows the input images for each task: MRI (task 1) or CBCT (task 2); the second column is the ground truth CT, and the third column is the associated dilated body outline.}
  \label{fig:synthrad_examples}
\end{figure}

\paragraph{SynthRAD2023}  
SynthRAD2023 is a synthetic radiology dataset designed for cross-modal brain image reconstruction tasks. It contains paired Computed Tomography (CT) and Magnetic Resonance Imaging (MRI) volumes with various degrees of sparsity and anatomical variability. The dataset simulates realistic clinical conditions by incorporating sparse CT slices and multiple MRI modalities, facilitating the evaluation of reconstruction methods under challenging sparse input scenarios.
Figure~\ref{fig:synthrad_examples} shows example images from the dataset. The images are reproduced from the original paper.

\paragraph{BraTS}  
The Brain Tumor Segmentation (BraTS) \cite{Menze_2015_BraTS} dataset is a widely used benchmark for brain tumor imaging studies. It includes multi-modal MRI scans (such as T1, T1c, T2, and FLAIR) from patients with gliomas, along with expert annotations. BraTS provides high-resolution volumetric MRI data with diverse tumor appearances, making it suitable for evaluating cross-modal synthesis and reconstruction algorithms in neuro-oncology. In our work, we utilize BraTS to validate the robustness of our method on real clinical MRI data with complex pathological structures.
Figure~\ref{fig:brats_examples} shows example images, reproduced from the original paper.

\section{Implementation Details}

Due to space limitations in the main paper, we present detailed training procedures in this section, with particular emphasis on slice selection and training logic.

We begin by training a BBDM \cite{Li_2023_BBDM} using paired CT and MRI data. This supervised training phase is relatively straightforward, with no additional data alignment or augmentation required beyond standard preprocessing.

Next, we construct a brain knowledge base by organizing the training data according to subject IDs. For datasets such as BraTS and SynthRAD2023, the original directory structures are inherently organized by subject, naturally forming an efficient retrieval database without the need for restructuring.

To enable retrieval-based synthesis, we fine-tune a retrieval model solely on the CT modality. We explored the potential of using all four modalities in BraTS—namely T1, T1ce (T1 with contrast enhancement), T2, and FLAIR—to improve generalization of the retrieval model. However, empirical results show that this multimodal input often degrades performance. We hypothesize this is due to the monotonous nature of grayscale medical images, where increasing the number of input channels introduces noise rather than useful diversity.

For training the ControlNet, we follow a retrieval-augmented setting. Given a source CT image \( y \), we retrieve another CT image \( r \) using the retrieval model. Our objective is to reconstruct the corresponding MRI of \( r \), using \( y \) as the input to the frozen BBDM and \( r \) as the conditioning input to ControlNet.
Only the ControlNet requires training in this phase. This significantly reduces training time compared to the original BBDM, as the parameter size of ControlNet is approximately half that of BBDM. The training process mimics a multi-agent setup: ControlNet is entirely dependent on the quality of the retrieval results. If the retrieval model performs poorly or the knowledge base lacks similar examples, ControlNet training becomes unstable. To mitigate this issue, we optionally apply spherical linear interpolation (slerp) between the retrieved image \( r \) and the input \( y \), generating additional intermediate samples to enhance training stability.

Joint training of BBDM and ControlNet presents further challenges. In particular, we observe significant loss fluctuations, especially when the retrieved image \( r \) differs substantially from the input \( y \). In such cases, the reconstructed MRI often lacks fine anatomical details.
We refer to related work such as EBDM \cite{Lee_2025_EBDM}, and conclude that the bottleneck primarily lies in data availability and retrieval quality.

In most cases, when \( y \) and \( r \) are reasonably similar, the resulting output exhibits the grayscale style of \( y \) and the structural content of \( r \). Occasionally, the output becomes nearly indistinguishable from \( r \), with minimal perceptual differences.

As discussed in the main paper, these limitations primarily stem from the lack of sufficiently similar samples in the retrieval database. Nevertheless, our method consistently outperforms baselines in terms of structural fidelity and style preservation, as evidenced by SSIM and PSNR scores.

\section{Unitized vs. Non-Unitized Objectives}

\subsection{Theoretical Analysis of Objective Functions}

\paragraph{Setup.}  
Let \(x_0\) denote the reference image and \(y\) the conditional image. Define
\begin{equation}
r = \|y - x_0\| \ge 0, \quad u = \frac{y - x_0}{\|y - x_0\|}, \quad y-x_0 = r u.
\end{equation}
Fix timestep \(t\) and write \(m_t\), \(\delta_t\); set \(s = \sqrt{\delta_t}\). Let \(\epsilon\sim \mathcal N(0,I)\) be independent Gaussian noise. Denote the model prediction as \(f\).

\begin{equation}
\text{Raw objective: } L_{\mathrm{raw}}(f) = \mathbb{E}\big[\|m_t r u + s \epsilon - f\|_2^2\big],
\end{equation}
\begin{equation}
\text{Unitized objective: } L_{\mathrm{unit}}(f) = \mathbb{E}\big[\|m_t u + s \epsilon - f\|_2^2\big].
\end{equation}

\noindent\textbf{Proposition 1.}  
The minimizers satisfy
\begin{equation}
f_{\mathrm{raw}}^\star = m_t \mathbb{E}[r u \mid \mathcal I], \quad 
f_{\mathrm{unit}}^\star = m_t \mathbb{E}[u \mid \mathcal I].
\end{equation}
In general \(f_{\mathrm{raw}}^\star \neq f_{\mathrm{unit}}^\star\).

\begin{proof}
For \(\ell_2\) loss, the minimizer is the conditional expectation:
\begin{equation}
f_{\mathrm{raw}}^\star = \mathbb{E}[m_t r u + s \epsilon \mid \mathcal I] = m_t \mathbb{E}[r u \mid \mathcal I],
\end{equation}
\begin{equation}
f_{\mathrm{unit}}^\star = m_t \mathbb{E}[u \mid \mathcal I].
\end{equation}
\end{proof}

\noindent\textbf{Proposition 2.}  
Let 
\begin{equation}
T_{\mathrm{raw}} = m_t r u + s \epsilon, \quad 
T_{\mathrm{unit}} = m_t u + s \epsilon.
\end{equation}
Gradients of squared error: \(-2(T-f)\). Covariance of gradient:
\begin{equation}
\mathrm{Cov}_{\mathrm{raw}} = 4\,\mathrm{Cov}(T_{\mathrm{raw}}) = 4 m_t^2 \mathrm{Cov}(r u) + 4 s^2 I,
\end{equation}
\begin{equation}
\mathrm{Cov}_{\mathrm{unit}} = 4\,\mathrm{Cov}(T_{\mathrm{unit}}) = 4 m_t^2 \mathrm{Cov}(u) + 4 s^2 I.
\end{equation}
If \(r\) varies across samples, \(\mathrm{Cov}(r u) \gg \mathrm{Cov}(u)\), increasing gradient variance.

\begin{proof}[Sketch]
Direct computation: for one sample \(-2(T-f)\), variance of stochastic gradient is
\(\mathbb{E}[(T-\mathbb{E}T)(T-\mathbb{E}T)^\top] = \mathrm{Cov}(T)\). Scaling by \(r\) inflates covariance proportionally to \(\mathrm{Var}(r)\).
\end{proof}

\bigskip
\noindent\textbf{Corollary.}  
For small training set \(n\), sample variance of \(r\) can deviate from population variance:
\begin{equation}
\mathrm{Var}_{\mathrm{emp}}(r) \sim \mathcal O(1/n),
\end{equation}
which amplifies stochastic gradient noise for raw objective. Unitization removes multiplicative scale, reducing gradient variance and improving stability.

\begin{proof}[Remarks]
Standard SGD bounds: larger \(\sigma^2 = \mathrm{Var}(\text{grad})\) implies slower convergence. Raw objective has extra \(\mathrm{Var}(r u)\) term; small datasets exacerbate instability.
\end{proof}

\subsection{Empirical Evaluation}

To verify the theoretical predictions, we conduct experiments comparing raw and unitized objectives on two representative datasets: BraTS and SynthRAD2023. While the convergence difference on BraTS is modest, the benefit of the unitized objective is clearly observed on SynthRAD2023, where BBDM converges substantially faster.

\begin{table*}[t]
\centering
\scriptsize
\begin{tabular}{lcccccccc}
\toprule
\multirow{2}{*}{Method} & \multicolumn{4}{c}{SynthRAD2023} & \multicolumn{4}{c}{BraTS} \\
\cmidrule(lr){2-5} \cmidrule(lr){6-9}
& NRMSE$\downarrow$ & PSNR$\uparrow$ & SSIM$\uparrow$ & I-SSIM$\uparrow$
& NRMSE$\downarrow$ & PSNR$\uparrow$ & SSIM$\uparrow$ & I-SSIM$\uparrow$ \\
\midrule
MaskGAN      & 0.1526$\pm$0.008 & 17.6850$\pm$0.032 & 0.5883$\pm$0.027 & 0.9093$\pm$0.014 & 0.1035$\pm$0.003 & 20.1710$\pm$0.045 & 0.7028$\pm$0.016 & 0.9215$\pm$0.008 \\
CT2MR        & 0.1138$\pm$0.002 & 19.2930$\pm$0.041 & 0.6015$\pm$0.015 & 0.9484$\pm$0.003 & 0.0643$\pm$0.003 & 23.8440$\pm$0.047 & 0.8535$\pm$0.007 & 0.9489$\pm$0.003 \\
Dual.        & 0.1093$\pm$0.003 & 18.1780$\pm$0.037 & 0.6162$\pm$0.006 & 0.9382$\pm$0.004 & 0.0706$\pm$0.002 & 24.5100$\pm$0.042 & 0.8419$\pm$0.006 & 0.9404$\pm$0.004 \\
I3Net        & 0.1653$\pm$0.014 & 11.7750$\pm$0.044 & 0.5581$\pm$0.038 & 0.9407$\pm$0.024 & 0.1502$\pm$0.010 & 18.9140$\pm$0.039 & 0.8074$\pm$0.019 & 0.9530$\pm$0.003 \\
ALDM         & 0.1336$\pm$0.009 & 17.5610$\pm$0.046 & 0.6266$\pm$0.015 & 0.9368$\pm$0.013 & 0.0808$\pm$0.009 & 21.9210$\pm$0.043 & 0.7930$\pm$0.011 & 0.9544$\pm$0.003 \\
mDAUNet   & 0.1069$\pm$0.012 & 18.6620$\pm$0.038 & 0.6331$\pm$0.008 & 0.9466$\pm$0.004 & 0.0602$\pm$0.004 & 24.6460$\pm$0.045 & 0.8378$\pm$0.009 & \textbf{0.9597$\pm$0.003} \\
\textbf{ReBrain} & \textbf{0.1054$\pm$0.002} & \textbf{20.7590$\pm$0.045} & \textbf{0.6682$\pm$0.009} & \textbf{0.9662$\pm$0.003} & \textbf{0.0551$\pm$0.002} & \textbf{25.6070$\pm$0.042} & \textbf{0.8887$\pm$0.008} & 0.9575$\pm$0.003 \\
\bottomrule
\end{tabular}
\caption{Quantitative comparison under half-resolution input on SynthRAD2023 and BraTS datasets, with mean ± standard deviation over repeated runs.}
\label{tab:quantitative_comparison_more}
\end{table*}

\begin{table}[t]
\centering
\renewcommand{\arraystretch}{1.0}
\setlength{\tabcolsep}{2pt}
\begin{tabular}{lcc}
\toprule
Dataset & Raw Objective (h) & Unitized Objective (h) \\
\midrule
BraTS         & 9.87 & \textbf{9.42} \\
SynthRAD2023  & 8.15 & \textbf{4.30} \\
\bottomrule
\end{tabular}
\caption{Training time comparison between raw and unitized objectives under multi-GPU parallel acceleration on BraTS and SynthRAD2023 datasets.}
\label{tab:training_time}
\end{table}

\begin{figure}[t]
    \centering
    \includegraphics[width=0.9\linewidth]{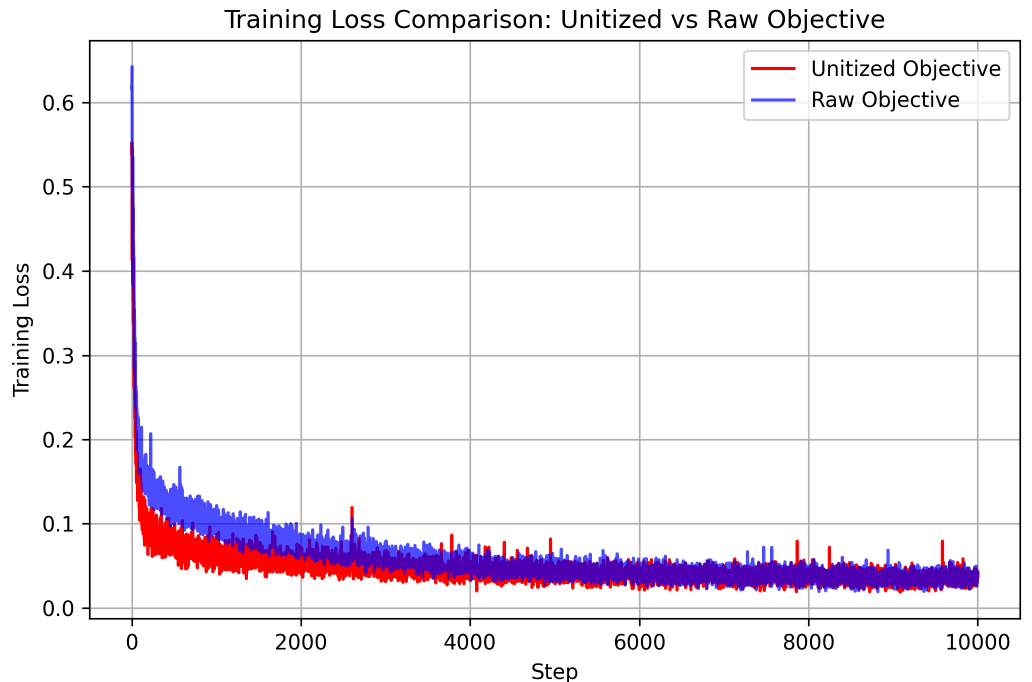}
    \caption{Training loss curves for raw and unitized objectives on SynthRAD2023. Unitized objective shows faster and smoother convergence on SynthRAD2023.}
    \label{fig:loss_curves}
\end{figure}

The loss curves in Figure~\ref{fig:loss_curves} further illustrate the effect: unitization leads to smoother and more stable convergence on SynthRAD2023.

Overall, we observed this effect somewhat incidentally: during training on SynthRAD2023, convergence was relatively difficult, and inspecting the loss curves revealed the advantage of the unitized objective. We hypothesize that the stronger benefit on SynthRAD2023 stems from its smaller dataset and higher variability between samples, which makes direct optimization of \(y - x_0\) more challenging. In contrast, BraTS contains more abundant and consistent data, leading to stable convergence even with the raw objective. In other words, when the sample set is small or heterogeneous, the scale of \(y - x_0\) can vary considerably across examples, increasing gradient variance and slowing convergence. Unitization mitigates this effect by removing the multiplicative scale, resulting in more stable and faster training. For large and more homogeneous datasets like BraTS, the variability is naturally lower, so both objectives perform similarly. Overall, unitization proves effective for improving training stability and speed, especially in datasets with limited size or high sample variability.

\section{More Results}

Table~\ref{tab:quantitative_comparison_more} presents a quantitative comparison under half-resolution input on the SynthRAD2023 and BraTS datasets. Across all metrics, ReBrain consistently achieves state-of-the-art performance, outperforming existing methods such as MaskGAN, CT2MR, Dual, I3Net, ALDM, and mDAUNet. Most of the reported standard deviations are small, indicating that the results are stable with low fluctuation. These results highlight the effectiveness and reliability of ReBrain in producing high-fidelity reconstructions under sparse input conditions.

\begin{figure}[t]
    \centering
    \includegraphics[width=0.48\textwidth]{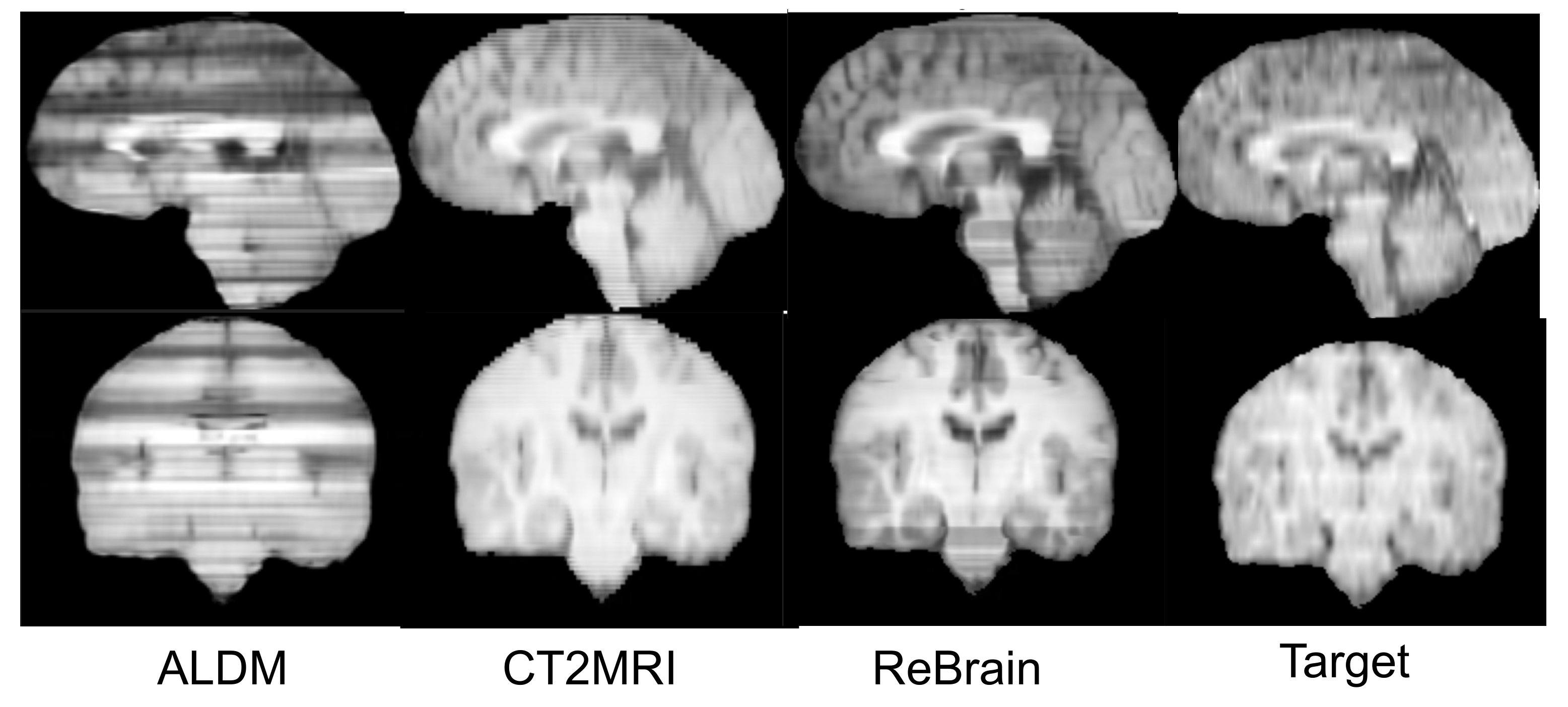}
    \caption{Comparison of generated results including ALDM, CT2MRI, and our ReBrain. ReBrain demonstrates superior fidelity and structural consistency.}
    \label{fig:extra_generation}
\end{figure}

Besides the results shown in Figure 4, we additionally provide the comparison including ALDM and CT2MRI. The generated results are visualized in Figure~\ref{fig:extra_generation}, highlighting that our ReBrain method produces superior fidelity and structural consistency.

\begin{figure}[t]
  \centering
  \includegraphics[width=1.0\linewidth]{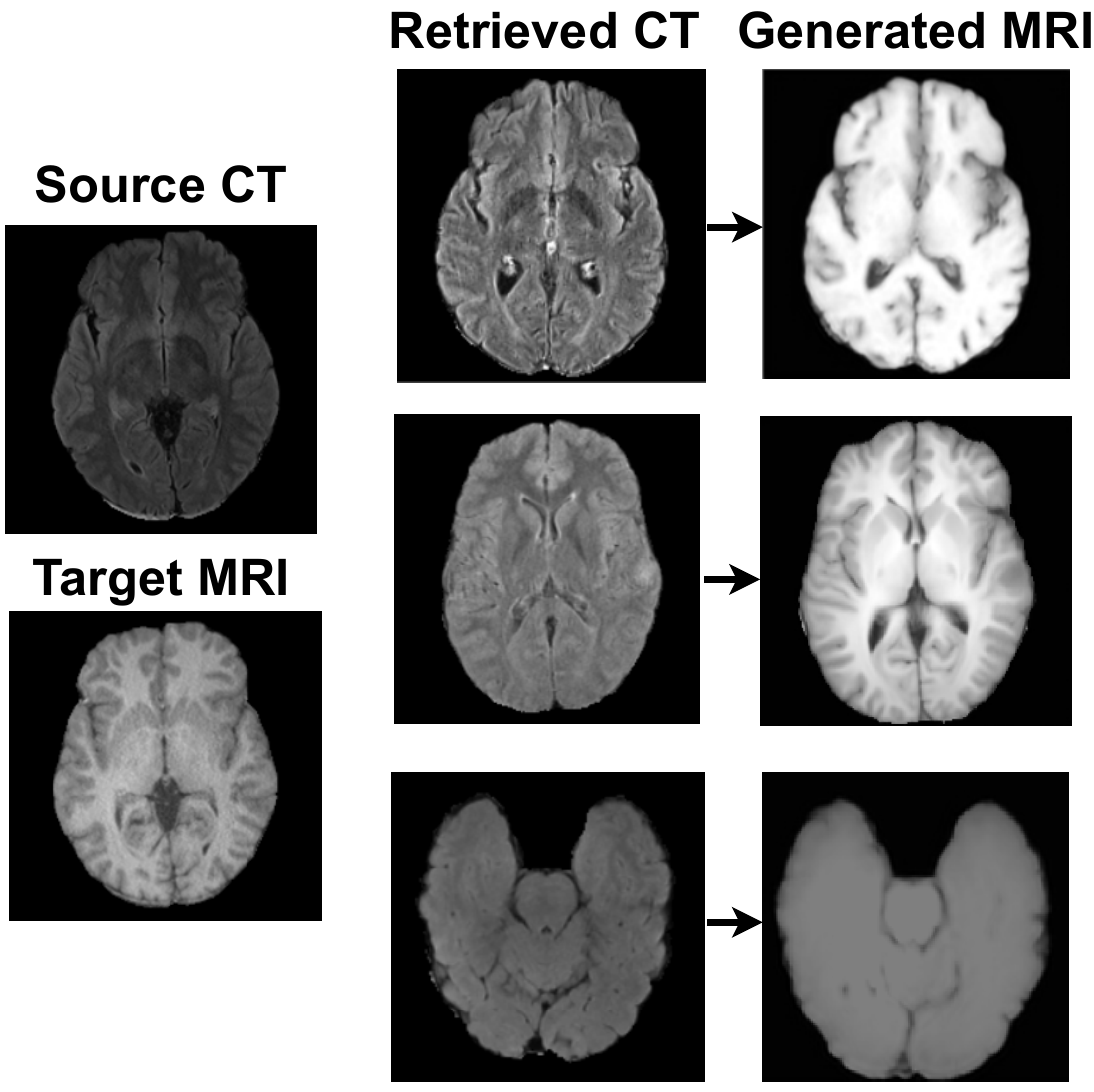}
  \caption{
  Illustration of uncertainty in generation caused by low similarity between input slice \( y \) and retrieved slice \( r \)
  }
  \label{fig:y_r}
\end{figure}

\textbf{Impact of Similarity Between \( y \) and \( r \).}  
In our framework, the retrieved slice \( r \) is used to guide the generation conditioned on the input slice \( y \). Typically, \( y \) and \( r \) are similar, enabling smooth and progressive synthesis. However, when the gap between \( y \) and \( r \) is large, the generation becomes highly stochastic, resulting in abrupt transitions rather than gradual interpolation.
As shown in Figure~\ref{fig:y_r}, we visualize multiple samples generated from a single \( y \) when paired with a poorly matched \( r \). This highlights an interesting uncertainty phenomenon. Considering the deterministic nature of medical imaging and the need to minimize risk, we employ interpolation instead of retrieval when the similarity between \( y \) and \( r \) is low.

\textbf{Effect of Using Standard Diffusion with ControlNet.} 
If we replace BBDM with a standard diffusion model while retaining ControlNet, the generated MRI tends to closely resemble \( r \), with limited incorporation of \( y \). This behavior is similar to directly feeding \( r \) into BBDM without ControlNet.
When the discrepancy between the source slice \( y \) and the retrieved slice \( r \) is large, we consider the standard diffusion process starting from Gaussian noise a viable alternative. In such cases, sampling from noise yields more reliable reconstructions than initializing from \( y \), which may impose incompatible semantic constraints on the generation.

\section{Limitations}

\textbf{Technical considerations.}  
We acknowledge that even when using \( y \) as the input to BBDM under the control of ControlNet, the generated structure still tends to approximate \( r \). Although this approach improves upon directly using \( r \) as the input to BBDM, there remains room for further enhancement. This outcome does not fully align with our expectation of a combined effect of \( y + r \).
We speculate that this is related to the ControlNet training process, where the target MRI paired with inputs \( y \) and \( r \) is predominantly based on \( r \).
In future work, we will continue to optimize our method guided by these insights.

In some cases, the retrieved slice \(r\) has high similarity with the target slice \(y\), but comes from a very different location in the brain. This causes a mismatch between appearance and true position, which may harm the quality of reconstruction. To reduce this problem, we added a simple check during retrieval: we remove slices whose positions differ too much from the query. This improves the trust in the results to some extent.

\textbf{Ethical and practical considerations.}  
Our framework uses previously observed, structurally similar slices only as a conditional reference to guide MRI reconstruction, rather than directly copying prior patient data. This ensures that retrieved slices act as a "soft hint," not a precise replication. While maintaining reasonable 3D structural continuity is one of the aims, it is achieved as part of balancing multiple reconstruction objectives rather than being the sole focus. SLERP interpolation is applied when retrieval fails or yields low-similarity slices, serving as a safe fallback that minimizes hallucinated structures.

Importantly, in clinical scenarios where the original CT lacks complete pathological information, no reconstruction method—including ours—can invent missing lesions. If a retrieved reference contains relevant pathology, it can assist in highlighting critical structures, potentially benefiting the patient. Conversely, if the patient is healthy or the reference has no pathology, any introduced errors are minor and limited to a careful review. Overall, this strategy substantially reduces hallucinations while keeping clinical risk very low, with potential benefits outweighing the unlikely costs.

To explicitly formalize the ethical trade-off between different types of errors, we define a weighted risk $R$ as follows:

\begin{equation}
R = w_\text{FN} \cdot \mathbf{1}[y=1 \wedge \hat{y}=0] + w_\text{FP} \cdot \mathbf{1}[y=0 \wedge \hat{y}=1],
\end{equation}

where $y \in \{0,1\}$ denotes the true pathology label ($1$ = lesion, $0$ = healthy), $\hat{y} \in \{0,1\}$ the predicted label from reconstructed slices, $w_\text{FN}$ is the weight for false negatives, and $w_\text{FP}$ is the weight for false positives. We set $w_\text{FN} \gg w_\text{FP}$ to reflect that missing a lesion is clinically far more severe than mistakenly flagging a healthy region.

Without retrieval, sparse or continuous 3D reconstruction methods may produce high $w_\text{FN}$ events, i.e., missing critical lesions. Our retrieval-augmented approach primarily shifts potential errors to $w_\text{FP}$-type events, which are less severe and can be mitigated through clinical review. This ensures that the most critical errors are minimized, while overall fidelity and structural consistency remain high, reducing potential harm to patients.

Moreover, we ensure that all reference slices used for guidance are fully anonymized, and no personally identifiable information is ever exposed. The retrieval mechanism is designed to prevent over-reliance on any single patient's data, further mitigating privacy risks. From an ethical standpoint, the model's predictions are intended as an assistive tool rather than a standalone diagnostic; final clinical decisions remain under the supervision of qualified professionals. Additionally, the model outputs are accompanied by uncertainty estimates and visual cues, promoting interpretability and enabling clinicians to verify reconstruction plausibility. These design choices collectively aim to respect patient privacy, reduce potential misuse, and maintain high standards of clinical responsibility.

\section{Clarification}

For clarity, we acknowledge an inconsistency in some of the illustrative figures: representative slice images from an MRI scan were inadvertently used to depict CT data modality. We emphasize that this minor presentational discrepancy does not affect the described methodology, algorithms, or any of the experimental results reported in this work. We clarify this here to ensure the utmost accuracy in the manuscript's visual documentation.

\begin{figure*}[t]
  \centering
  \includegraphics[width=\linewidth]{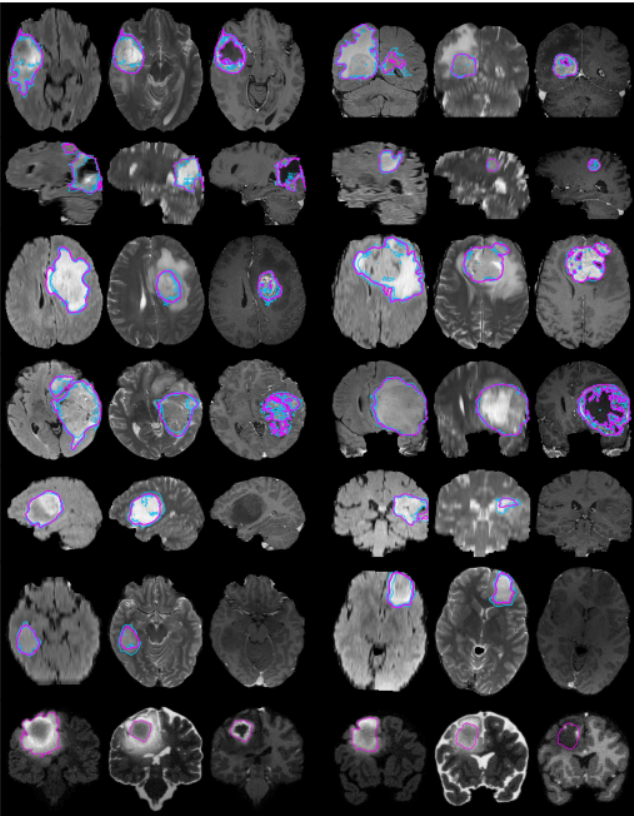}
  \caption{Examples from the BraTS training data, with tumor regions as inferred from the annotations of individual experts (blue lines) and consensus segmentation (magenta lines). Each row shows two cases of high-grade tumor (rows 1--4), low-grade tumor (rows 5--6), or synthetic cases (last row). Images vary between axial, sagittal, and transversal views, showing for each case: FLAIR with outlines of the whole tumor region (left); T2 with outlines of the core region (center); T1c with outlines of the active tumor region if present (right).}
  \label{fig:brats_examples}
\end{figure*}

\end{document}